\documentclass[letterpaper]{article} 
\usepackage[preprint]{aaai2027}  
\usepackage[hyphens]{url}  
\usepackage{graphicx} 
\usepackage{natbib}  
\usepackage{caption} 
\usepackage{algorithm}
\usepackage{float}
\usepackage{amsmath}
\usepackage{amssymb}
\usepackage{algorithm}
\usepackage{algpseudocode}
\usepackage{booktabs}
\usepackage{newfloat}
\usepackage{listings}
\DeclareCaptionStyle{ruled}{labelfont=normalfont,labelsep=colon,strut=off} 
\floatstyle{ruled}
\newfloat{listing}{tb}{lst}{}
\floatname{listing}{Listing}

\usepackage{booktabs}

\title{Multiphase-Diff: Diffusion-Based Generative Modeling for High-Contrast Multiphase Physical Systems with Sharp Interfaces}

\author{
    Yining Huang\textsuperscript{\rm 1},
    Zhenyu Liang\textsuperscript{\rm 2,3}\corresponding
}
\affiliations{
    \textsuperscript{\rm 1}University of Texas at Austin\\
    \textsuperscript{\rm 2}The Hong Kong University of Science and Technology\\
    \textsuperscript{\rm 3}University of Maryland, College Park\\

}

\newcommand{\E}{\mathbb{E}}
\newcommand{\Normal}{\mathcal{N}}
\newcommand{\Rtilde}{\widetilde{R}}
\newcommand{\abar}{\bar{\alpha}}
\newcommand{\ind}{\mathbf{1}}
\newcommand{\Wone}{W_1}

\DeclareMathOperator{\median}{median}
\DeclareMathOperator{\stopgrad}{stopgrad}

\begin{document}

\maketitle

\begin{abstract}
Physics-constrained diffusion for high-contrast, sharp-interface multiphase fields faces three coupled difficulties. At coefficient jumps, expanded pointwise strong-form PDE residuals contain singular gradient terms that can penalize physical interfaces. Under extreme contrast, low-magnitude phases may fall below the diffusion noise floor and be erased, misscaled, or generated with negative coefficients, while a global likelihood scale allows high-magnitude phases to dominate supervision. We therefore propose \textbf{Multiphase-Diff}, which makes three corresponding contributions: (i) a conservative flux residual that avoids differentiating discontinuous coefficients and enforces discrete conservation; (ii) an analytic bijective representation that maps low-amplitude signals to order-one latent scales and guarantees coefficient positivity through exponential decoding; and (iii) a Jacobi-preconditioned likelihood that normalizes local residual scales for balanced supervision. Experiments on three complementary multiphase benchmarks demonstrate the superiority of Multiphase-Diff over seven baselines in both physical and distributional fidelity and its robustness across phase contrasts and compositions, establishing its effectiveness for scientific sample generation in this challenging regime.
\end{abstract}

\section{Introduction}
\label{sec:introduction}

Learning generative models that capture the structure of complex scientific data while respecting known physical principles is a central objective of AI for science ~\cite{karniadakis2021physics}. In scientific domains, fidelity extends beyond statistical agreement with observed data: generated physical fields must also satisfy the physical admissibility requirements imposed by conservation laws, constitutive relations, and boundary and interface conditions. This requirement is particularly challenging in multiphase systems with extreme material contrasts and sharp interfaces, where heterogeneous statistics interact with nonsmooth physics.

Denoising diffusion probabilistic models (DDPMs)~\cite{ho2020ddpm,sohldickstein2015nonequilibrium} have attracted widespread attention in generative modeling because of their training stability, broad mode coverage, and high sample quality~\cite{nichol2021improved,song2021scorebased}. However, as purely data-driven models, their denoising objective learns to reproduce the statistical patterns of the training data, thereby promoting {\it distributional fidelity} without necessarily ensuring {\it physical fidelity}. Under this paradigm, a generated field may appear plausible while violating the governing conservation laws, limiting its use in downstream scientific applications.

Recent studies incorporate partial differential equations (PDEs) into diffusion models to improve the physical consistency of generated fields. For example, residuals derived from the Darcy equation can guide fluid-field generation toward physically admissible solutions. Some of these methods have been evaluated primarily on smooth fields or systems with moderate heterogeneity~\cite{shu2023physics,jacobsen2025cocogen}. Moving beyond these relatively regular settings, DiffusionPDE~\cite{huang2024diffusionpde} considers binary Darcy coefficients but evaluates them only at moderate contrast. FunDiff~\cite{wang2026fundiff} considers Burgers solutions with sharp gradients and elasticity fields containing material interfaces, but does not study discontinuous material coefficients under extreme contrast. Taken together, the physics-informed diffusion-based generation of physical fields with high material contrast and sharp interfaces remains insufficiently studied.

High-contrast, sharp-interface multiphase fields pose three challenges: \textbf{Interface irregularity.} Expanded pointwise strong-form residuals are not classically well defined at coefficient jumps, creating artificial penalties that can blur or suppress interfaces. \textbf{Low-amplitude phase under-resolution.} Low-magnitude phases can have negligible signal-to-noise ratios during diffusion and be underweighted by absolute-error training, causing them to be erased, misscaled, or driven to inadmissible negative values. \textbf{Residual-scale heterogeneity.} Residual magnitudes vary substantially across phases and samples, causing a global likelihood scale to be dominated by high-flux regions and preventing balanced supervision.

To solve these problems, we propose \textbf{Multiphase-Diff}, whose contributions are as follows:

\begin{enumerate}
\item We introduce a \textbf{conservative flux residual} based on finite-volume two-point fluxes and harmonic face averaging, which avoids differentiating the coefficient, remains well posed across coefficient jumps of arbitrary contrast, and enforces discrete conservation.

\item We develop an \textbf{analytic bijective representation} using logarithmic and inverse hyperbolic sine transformations to resolve low-amplitude phases and ensure coefficient positivity, together with a \textbf{Jacobi-preconditioned residual likelihood} that accounts for local residual scales and balances physics supervision across phases and samples with widely varying magnitudes.

\item We validate Multiphase-Diff against a broad set of baselines on three complementary discontinuous multiphase benchmarks, demonstrating improved physical fidelity and robustness to phase contrast and composition.

\end{enumerate}

\section{Related Work}
\label{sec:related-work}

\subsection{Generative Modeling of Multiphase Systems}
Generative models are increasingly used to capture the complex geometries and statistical distributions of multiphase systems. Early studies employed GANs to reconstruct multiphase electrode microstructures~\cite{gayonlombardo2020pores}, variational autoencoders to explore the design space of dual-phase materials~\cite{attari2023inverse}, and likelihood-based generative flows to synthesize three-dimensional porous media~\cite{guan2021flow}. More recently, diffusion models have been applied to conditional reservoir-facies generation and facies-based geomodel parameterization~\cite{lee2025facies,difederico2025latent}, as well as to the reconstruction of heterogeneous and multiphase porous microstructures~\cite{lyu2024microstructure,zhu2025porescale,baishnab2025multiphase}. Conditional diffusion models further control effective properties such as permeability, phase volume fraction, and tortuosity~\cite{lyu2024microstructure,baishnab2025multiphase}, and related methods support property-driven microstructure design and structural topology optimization~\cite{vlassis2023microstructures,maze2023topology,giannone2023aligning}. Across this literature, the generated output is the phase morphology or material layout. Physical quantities serve mainly as conditioning variables, design objectives, or evaluation metrics rather than being generated jointly with the material field under a local conservation law. Consequently, these methods do not directly model coefficient--solution pairs constrained by local conservation.

\subsection{Physics-Constrained Diffusion Models}
Physics-constrained diffusion models introduce PDE information during either training or sampling. PIDM formulates the PDE residual as a training-time likelihood~\cite{bastek2025pidm}, while physics-informed consistency models apply residual supervision in a separate fine-tuning stage~\cite{chang2026stabilizing}. Gradient-based approaches use residual information to steer generation: PG-Diffusion conditions denoising on residual gradients~\cite{shu2023physics}, whereas CoCoGen and DiffusionPDE correct intermediate samples during inference~\cite{jacobsen2025cocogen,huang2024diffusionpde}. Related guidance has also been developed for scaled spectral latent representations~\cite{gallon2026spectral}. Beyond direct residual guidance, FunDiff incorporates physical priors into a function autoencoder before latent generation with rectified flow~\cite{wang2026fundiff}, and physics-informed distillation converts a pretrained diffusion model into a PDE-supervised one-step generator~\cite{zhang2025distillation}. Despite these advances, these methods are not designed for extreme-contrast discontinuities, which require the field representation, residual discretization, and likelihood scaling to accommodate sharp interfaces and widely separated phase magnitudes.

\section{Preliminary}
\label{sec:preliminary}

A denoising diffusion probabilistic model (DDPM)~\cite{ho2020ddpm} consists of a fixed forward process $q$ and a learned reverse process $p_\theta$. Let $x_0\sim q_{\mathrm{data}}$ be a clean sample and $x_t$ its noisy version at timestep $t\in\{1,\ldots,T\}$, where $T$ is the number of diffusion steps. We use $\mathcal N(x;m,C)$ for a Gaussian distribution over $x$ with mean $m$ and covariance $C$, and $I$ for the identity matrix.

\noindent\textbf{Forward process.}
Given a variance schedule $\{\beta_t\in(0,1)\}_{t=1}^{T}$, define $\alpha_t=1-\beta_t$, $\bar\alpha_t=\prod_{s=1}^{t}\alpha_s$, and $\bar\alpha_0=1$. The one-step transition and direct sampling formula are

\begin{equation}
\begin{gathered}
q(x_t\mid x_{t-1})
=\mathcal N\!\left(
x_t;\sqrt{\alpha_t}\,x_{t-1},\beta_t I
\right)\\
x_t
=\sqrt{\bar\alpha_t}\,x_0
+\sqrt{1-\bar\alpha_t}\,\epsilon
\qquad \epsilon\sim\mathcal N(0,I)
\end{gathered}
\label{eq:forward-process}
\end{equation}

The schedule is chosen so that $\bar\alpha_T\approx0$, making $x_T$ approximately standard Gaussian noise.

\noindent\textbf{Reverse process.}
Generation starts from $p(x_T)=\mathcal N(x_T;0,I)$. Under the $x_0$-prediction parameterization, the network $f_\theta$ with parameters $\theta$ predicts $\hat x_0=f_\theta(x_t,t)$. Substituting $\hat x_0$ into the forward posterior gives the reverse mean, while the scalar variance is fixed to its posterior value:

\begin{equation}
\begin{gathered}
p_\theta(x_{t-1}\mid x_t)
=\mathcal N\!\left(x_{t-1};\mu_{\theta,t},\Sigma_t I\right)\\
\mu_{\theta,t}
=\frac{\sqrt{\bar\alpha_{t-1}}\,\beta_t}
       {1-\bar\alpha_t}\,\hat x_0
+\frac{\sqrt{\alpha_t}(1-\bar\alpha_{t-1})}
       {1-\bar\alpha_t}\,x_t\\
\Sigma_t
=\frac{1-\bar\alpha_{t-1}}
       {1-\bar\alpha_t}\,\beta_t
\end{gathered}
\label{eq:reverse-process}
\end{equation}

Applying these transitions from $t=T$ to $t=1$ transforms Gaussian noise into a sample from the learned distribution. The training objective is introduced in Section~\ref{sec:training-objective}; further details appear in the original DDPM paper~\cite{ho2020ddpm}.

\begin{figure*}[t]
    \centering
    \includegraphics[width=\textwidth]{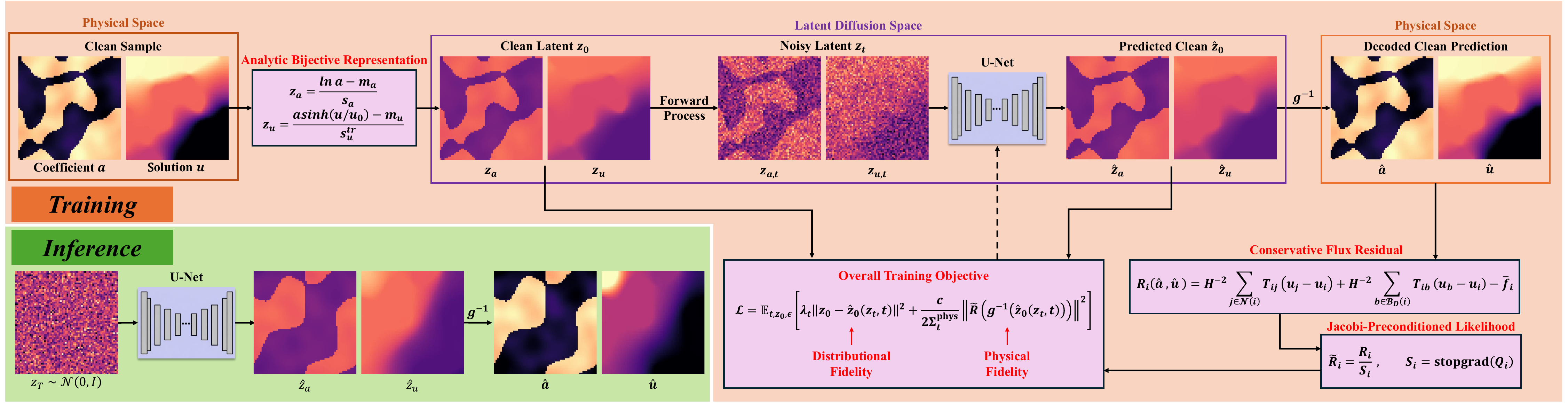}
    \caption{Overview of Multiphase-Diff, illustrated with a Darcy coefficient--solution pair. During training, the analytic bijection $g$ maps $(a,u)$ to the clean latent $z_0$, which is corrupted by the forward process and reconstructed by a clean-latent-predicting U-Net. The prediction $\hat z_0$ is decoded to $(\hat a,\hat u)$, whose conservative finite-volume flux residual is Jacobi-preconditioned to define the physics loss combined with the latent data loss. These mechanisms keep low-amplitude phases resolvable and coefficients positive, enforce interface-compatible local conservation, and balance residual scales across phases and samples, respectively. Inference requires no residual correction and performs standard reverse diffusion in latent space followed by one decoding step.}
    \label{fig:framework}
\end{figure*}

\section{Methodology}
\label{sec:methodology}

We propose \textbf{Multiphase-Diff}, a physics-constrained diffusion framework for generative modeling of high-contrast multiphase fields with sharp interfaces. As illustrated in Figure~\ref{fig:framework}, it learns the joint distribution of a positive material coefficient $a(\mathbf{x})$ and its solution $u(\mathbf{x})$, governed by $\nabla\cdot(a(\mathbf{x})\nabla u(\mathbf{x}))=f(\mathbf{x})$ for $\mathbf{x}\in\Omega$ with prescribed boundary conditions. Here, $a$ denotes a material property such as permeability, viscosity, or conductivity, $u$ is the corresponding physical response, and $f$ is a prescribed source. Given paired observations of $(a,u)$, Multiphase-Diff generates coefficient--solution pairs that reproduce the data distribution while satisfying local conservation.

\subsection{Conservative Flux Residual}
\label{sec:flux-residual}

The principal difficulty arises at material interfaces where $a$ is discontinuous. Consider a positive piecewise-constant coefficient on $\Omega\subset\mathbb R^2$. Across a perfect interface without interfacial sources or contact resistance, the solution and normal flux satisfy

\begin{equation}
[\,u\,]=0
\qquad
[\,a\,\partial_n u\,]=0
\label{eq:interface-conditions}
\end{equation}
where $[\cdot]$ denotes the interface jump and $n$ its unit normal. Consequently, the normal derivative generally scales inversely with the coefficient. Some physics-constrained diffusion methods evaluate PDE residuals pointwise on predicted fields~\cite{bastek2025pidm,jacobsen2025cocogen,huang2024diffusionpde}, which is effective for smooth fields but can introduce a nonphysical signal at discontinuous leading coefficients. This limitation is evident from the expanded strong form

\begin{equation}
\nabla\cdot(a\nabla u)
=a\,\Delta u+\nabla a\cdot\nabla u
\label{eq:expanded-strong-form}
\end{equation}

Because $a$ is piecewise constant, $\nabla a$ contains a singular interface contribution, and $\nabla a\cdot\nabla u$ is not classically defined there. On a grid with spacing $H$, directly approximating $\nabla a$ at an interface scales as $\Delta a/H$, where $\Delta a$ is the coefficient jump. The resulting artificial residual grows with contrast and grid refinement, penalizing the interface itself and potentially encouraging interface blurring or phase suppression.

We instead introduce a conservative flux residual based on discrete cellwise balance. At each sampled diffusion time, the predicted coefficient--solution pair is decoded into physical space and evaluated using differentiable two-point fluxes with harmonic transmissibilities, following locally conservative finite-volume principles~\cite{aavatsmark2002multipoint,edwards1998finite}. This construction enforces flux continuity without differentiating or penalizing the coefficient jump.

On the uniform orthogonal cell-centered grid used here, let $a_i$ and $u_i$ denote the values in cell $i$, $\mathcal N(i)$ its neighbors, and $\mathcal B_D(i)$ and $\mathcal B_N(i)$ its Dirichlet and Neumann boundary faces. For $b\in\mathcal B_D(i)$, $u_b$ is prescribed; for $b\in\mathcal B_N(i)$, $\phi_b=(a\nabla u)\cdot n_b$ is the prescribed outward flux. We absorb the Neumann contribution into the boundary-adjusted source
$\bar f_i=f_i-H^{-1}\sum_{b\in\mathcal B_N(i)}\phi_b$,
where $f_i$ is the cell-averaged source. The residual is

\begin{equation}
\begin{aligned}
R_i(a,u)
&=H^{-2}\sum_{j\in\mathcal N(i)}
T_{ij}(u_j-u_i)\\
&\quad+H^{-2}\sum_{b\in\mathcal B_D(i)}
T_{ib}(u_b-u_i)-\bar f_i
\end{aligned}
\label{eq:conservative-residual}
\end{equation}

The transmissibilities are

\begin{equation}
T_{ij}=\frac{2a_i a_j}{a_i+a_j}
\qquad
T_{ib}=2a_i
\label{eq:transmissibility}
\end{equation}

This construction is suited to grid-resolved sharp interfaces because \textbf{(i)} the coefficient enters through harmonic face averaging, which remains finite at any positive contrast and limits the flux according to the smaller adjacent coefficient without introducing a singular interface term; \textbf{(ii)} each internal face flux enters adjacent cell balances with opposite signs and cancels exactly, ensuring discrete conservation; and \textbf{(iii)} the residual uses the data-generating discretization, allowing reference fields to approach solver tolerance without discretization mismatch. The resulting transmissibilities require positive coefficients, motivating the positivity-preserving representation introduced next.

\subsection{Analytic Bijective Representation}
\label{sec:representation}

Although the conservative residual resolves the interface singularity, high contrast remains difficult for diffusion. Normalizing the high phase to $a_H=1$ with contrast $\gamma=10^6$ places the low phase at $a_L=10^{-6}$. At a timestep with noise standard deviation $10^{-2}$, this signal is four orders of magnitude below the noise. More generally, $\operatorname{SNR}_{L}(t)=\bar\alpha_t a_L^2/(1-\bar\alpha_t)$. This SNR and the squared-loss contribution under equal relative error are $\gamma^{-2}$ times those of the high phase, or $10^{-12}$ in this example. The model may therefore erase or misscale the low phase, and small denoising errors may drive a positive coefficient below zero.

We perform diffusion in the analytic bijective space $z_0=g(a,u)=(z_a,z_u)$:

\begin{equation}
z_a=\frac{\ln a-m_a}{s_a}
\qquad
z_u=\frac{\operatorname{asinh}(u/u_0)-m_u}{s_u^{\mathrm{tr}}}
\label{eq:bijective-representation}
\end{equation}

Here, $m_a,s_a$ and $m_u,s_u^{\mathrm{tr}}$ are the training-set means and standard deviations of $\ln a$ and $\operatorname{asinh}(u/u_0)$, respectively, and $u_0>0$ is a fixed solution scale. The logarithm converts multiplicative contrast into additive separation: $a_H=1$ and $a_L=10^{-6}$ become $0$ and $-13.8$ before standardization and approximately $+1$ and $-1$ for balanced phases afterward. Diffusion thus operates on separated order-one modes, while logarithmic differences represent relative coefficient errors.

Decoding gives $a=\exp(m_a+s_a z_a)>0$ and $u=u_0\sinh(m_u+s_u^{\mathrm{tr}}z_u)$. Hence, coefficients used for generation and physics evaluation remain positive. The inverse hyperbolic sine transformation compresses signed solutions while remaining linear near zero. Unlike learned latent autoencoders~\cite{rombach2022latent}, both maps are non-learned, invertible, and monotone. At each timestep, the predicted clean latent $\hat z_0$ is decoded as $(\hat a,\hat u)=g^{-1}(\hat z_0)$ before evaluating physics. The residual scale can nevertheless vary across phases and samples, motivating the following normalization.

\subsection{Jacobi-Preconditioned Likelihood}
\label{sec:preconditioned-likelihood}


The preceding components address interface irregularity, low-phase resolution, and coefficient positivity. However, a common global scale for absolute residuals remains problematic. If phases with $a_H=1$ and $a_L=10^{-6}$ both contain a $1\%$ relative error, their absolute residuals may differ by approximately $10^6$, making their contributions to the squared loss differ by approximately $10^{12}$ when other factors are comparable. The optimizer is then driven primarily by high-phase residuals.

We convert each absolute residual into a relative physical violation. Define the unscaled diagonal coefficient of the flux stencil as $D_i=\sum_{j\in\mathcal N(i)}T_{ij}+\sum_{b\in\mathcal B_D(i)}T_{ib}$. It collects the transmissibilities connected to cell $i$ and measures the local sensitivity of its balance equation to changes in the solution. Accordingly, $D_i/H^2$ is the magnitude of the operator diagonal, and $D_i s_{\hat u}/H^2$ estimates the natural residual scale. Here, $s_{\hat u}>0$ is a detached robust per-sample scale of the decoded solution, with the solution gauge fixed in Neumann cases. We define $Q_i=D_i s_{\hat u}/H^2+\lVert f\rVert_\infty+\varepsilon$, where $\varepsilon$ is a positive numerical floor, and use

\begin{equation}
S_i=\operatorname{stopgrad}(Q_i)
\qquad
\widetilde R_i=\frac{R_i}{S_i}
\label{eq:jacobi-preconditioning}
\end{equation}

This Jacobi normalization rescales each equation by its local operator magnitude. It reduces phase- and sample-dependent residual-scale variation while preserving the conservation target because $\widetilde R_i=0$ if and only if $R_i=0$. Detaching $Q_i$ removes the direct gradient incentive to inflate predicted scales. Apart from $\varepsilon$, $\widetilde R$ is dimensionless and invariant under $(a,u,f)\mapsto(\lambda a,u,\lambda f)$. We denote this normalization by $\widetilde R=\mathcal P(a,u,f)$ and use it as the mean of a zero-residual likelihood below.

\subsection{Training Objective}
\label{sec:training-objective}

Applying the clean-sample parameterization of Section~\ref{sec:preliminary} in latent space, the network predicts $\hat z_0(z_t,t)$ from $z_t$ generated by~\eqref{eq:forward-process}. Its equivalence to noise prediction under timestep-dependent weighting is derived in the Supplementary Material. We apply the Min-SNR weight $\lambda_t$~\cite{hang2023minsnr} to latent regression and evaluate physics on $g^{-1}(\hat z_0)$.

Following the virtual-observation formulation~\cite{rixner2021virtual,bastek2025pidm}, physical consistency is represented by the zero-residual observation $\hat r=0$ with likelihood

\begin{equation}
q_R(\hat r=0\mid\hat z_0,t)
=\mathcal N\!\left(
0;\widetilde R\!\left(g^{-1}(\hat z_0)\right),
\frac{\Sigma_t^{\mathrm{phys}}}{c}I
\right)
\label{eq:residual-likelihood}
\end{equation}

where $c>0$ controls physics supervision and
$\Sigma_t^{\mathrm{phys}}=\max(\Sigma_t,\Sigma_{\min})$, with $\Sigma_{\min}>0$, prevents the likelihood variance from vanishing when $\Sigma_1=0$. Combining its negative log-likelihood with latent regression gives

\begin{equation}
\begin{aligned}
\mathcal L
=\mathbb E_{t,z_0,\epsilon}\Bigg[
&\lambda_t\lVert z_0-\hat z_0(z_t,t)\rVert^2\\
&+\frac{c}{2\Sigma_t^{\mathrm{phys}}}
\left\lVert
\widetilde R\!\left(g^{-1}(\hat z_0(z_t,t))\right)
\right\rVert^2
\Bigg]
\end{aligned}
\label{eq:training-objective}
\end{equation}

The terms promote latent distributional fidelity and decoded physical consistency, respectively. Training adds one decoding and residual evaluation per iteration, while inference uses standard latent diffusion followed by one decoding step. Algorithm~\ref{alg:multiphase-diff} summarizes both procedures.

\begin{algorithm}[t]
\caption{Multiphase-Diff Training and Sampling}
\label{alg:multiphase-diff}
\begin{algorithmic}[1]
\Require Data distribution $q_{\mathrm{data}}(a,u)$; bijection $g$
\Require Flux residual $R$; Jacobi normalization $\mathcal P$
\Require Model $f_\theta$; schedules
$\{\bar\alpha_t,\lambda_t,\Sigma_t\}_{t=1}^{T}$; $c,\Sigma_{\min}$
\Repeat
    \State Sample $(a,u)\sim q_{\mathrm{data}}(a,u)$
    \State $z_0\gets g(a,u)$
    \State Sample $t\sim\mathrm{Uniform}\{1,\ldots,T\}$ and
           $\epsilon\sim\mathcal N(0,I)$
    \State $z_t\gets\sqrt{\bar\alpha_t}\,z_0
           +\sqrt{1-\bar\alpha_t}\,\epsilon$
    \State $\hat z_0\gets f_\theta(z_t,t)$
    \State $\mathcal L_{\mathrm{data}}
           \gets\lambda_t\lVert z_0-\hat z_0\rVert^2$
    \State $(\hat a,\hat u)\gets g^{-1}(\hat z_0)$
    \State $\widetilde R\gets\mathcal P(\hat a,\hat u,f)$
    \State $\Sigma_t^{\mathrm{phys}}
           \gets\max(\Sigma_t,\Sigma_{\min})$
    \State $\mathcal L\gets\mathcal L_{\mathrm{data}}
           +\dfrac{c}{2\Sigma_t^{\mathrm{phys}}}
            \lVert\widetilde R\rVert^2$
    \State Update $\theta$ using $\nabla_\theta\mathcal L$
\Until{convergence}
\Statex
\Statex \textbf{Sampling}
\State Draw $z_T\sim\mathcal N(0,I)$ and apply the learned reverse
       process to obtain $\hat z_0$
\Return $(\hat a,\hat u)=g^{-1}(\hat z_0)$
\end{algorithmic}
\end{algorithm}

\begin{table*}[t]
\centering
\small
\setlength{\tabcolsep}{2.2pt}
\begin{tabular*}{\textwidth}{
@{\extracolsep{\fill}}c*{8}{c}@{}
}
\toprule
& DDPM
& PG-Diffusion
& CoCoGen
& DiffusionPDE
& FunDiff
& PIDM
& PIDM-log
& Multiphase-Diff \\
\midrule

\multicolumn{9}{c}{(a) Physical Residual Fidelity (PRF) $\downarrow$} \\
\midrule
Case 1 & $1.09\!\times\!10^{-2}$ & $1.34\!\times\!10^{-2}$ & $1.96\!\times\!10^{-2}$ & $9.76\!\times\!10^{-3}$ & $6.28\!\times\!10^{-3}$ & $9.81\!\times\!10^{-4}$ & $8.26\!\times\!10^{-4}$ & $\boldsymbol{3.58\!\times\!10^{-4}}$ \\
Case 2 & $8.57\!\times\!10^{-2}$ & $1.14\!\times\!10^{-1}$ & $1.60\!\times\!10^{-1}$ & $9.18\!\times\!10^{-2}$ & $1.47\!\times\!10^{-3}$ & $4.57\!\times\!10^{-1}$ & $1.61\!\times\!10^{-3}$ & $\boldsymbol{2.30\!\times\!10^{-4}}$ \\
Case 3 & $3.43\!\times\!10^{-3}$ & $3.16\!\times\!10^{-3}$ & $1.24\!\times\!10^{-3}$ & $3.23\!\times\!10^{-3}$ & $4.48\!\times\!10^{-3}$ & $1.68\!\times\!10^{-3}$ & $3.15\!\times\!10^{-3}$ & $\boldsymbol{4.17\!\times\!10^{-4}}$ \\

\midrule
\multicolumn{9}{c}{(b) Negative-coefficient rate (Neg, \%) $\downarrow$} \\
\midrule
Case 1 & 1.9 & 4.5 & 6.9 & 2.8 & \textbf{0.0} & 2.4 & \textbf{0.0} & \textbf{0.0} \\
Case 2 & 7.0 & 11.1 & 14.8 & 8.8 & \textbf{0.0} & 43.0 & \textbf{0.0} & \textbf{0.0} \\
Case 3 & 20.0 & 28.1 & 33.1 & 27.0 & \textbf{0.0} & 3.2 & \textbf{0.0} & \textbf{0.0} \\

\midrule
\multicolumn{9}{c}{(c) Worst per-phase Wasserstein-1 distance of $\ln a$ ($W_1$) $\downarrow$} \\
\midrule
Case 1 & 0.144 & 0.298 & 0.116 & 0.174 & 0.038 & 1.305 & 0.179 & \textbf{0.008} \\
Case 2 & 1.141 & 1.140 & 1.149 & 1.187 & 1.339 & 3.397 & 1.422 & \textbf{0.933} \\
Case 3 & 2.522 & 2.520 & 2.715 & 2.502 & 0.739 & 2.486 & 1.282 & \textbf{0.025} \\

\midrule
\multicolumn{9}{c}{(d) Interface sharpness (Sharp) $\downarrow$} \\
\midrule
Case 1 & 0.135 & 0.237 & 0.104 & 0.150 & 0.126 & 0.123 & 0.055 & \textbf{0.023} \\
Case 2 & 0.120 & 0.118 & 0.099 & 0.131 & 0.045 & 0.240 & 0.128 & \textbf{0.023} \\
Case 3 & 0.733 & 0.792 & 0.639 & 0.720 & 0.374 & 1.154 & 0.193 & \textbf{0.039} \\
\bottomrule
\end{tabular*}
\caption{Baseline comparison over 256 generated samples. Lower is better for all metrics, and the best values are shown in bold.}
\label{tab:baseline_comparison}
\end{table*}


\begin{figure}[H]
    \centering
    \includegraphics[
        width=\columnwidth,
        keepaspectratio
    ]{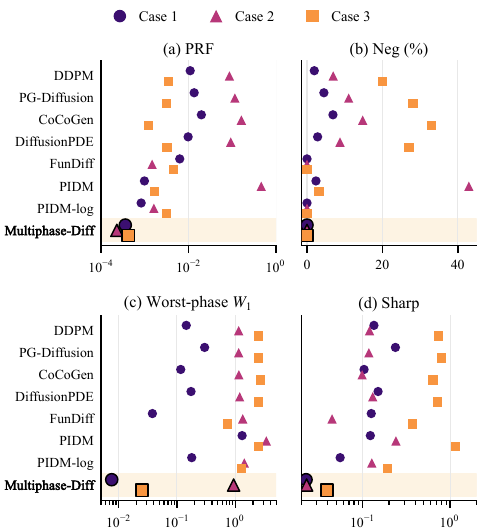}
    \caption{Four-metric visual comparison across three benchmarks. PRF, $W_1$, and Sharp use log scales; Neg is linear. Shading highlights Multiphase-Diff's superior performance.}
    \label{fig:metric_comparison}
\end{figure}

\section{Experiments and Results}
\label{sec:experiments}

We evaluate Multiphase-Diff through three complementary studies: Section~\ref{sec:baseline_comparison} compares it against seven baselines, Section~\ref{sec:ablations} ablates the three components, and Section~\ref{sec:robustness} evaluates its robustness to phase contrast and phase composition.

\subsection{Experimental Setup}
\label{sec:experimental_setup}

\paragraph{Datasets.} We consider three complementary multiphase benchmarks. Case~1 models Darcy flow through a two-facies porous medium and jointly generates the permeability $K$ and pressure $p$; $K$ has irregular sharp jumps with a contrast of approximately $10^{3}$, while $p$ remains continuous but develops discontinuous gradients across facies interfaces. Case~2 models fully developed gas--liquid duct flow and generates the viscosity $\mu$ and axial velocity $w$; the two phases produce sharp viscosity interfaces, contrasts from $55$ to $10^{4}$, and strongly heterogeneous velocity scales. Case~3 models electronic conduction through a three-phase electrode and generates the conductivity $\sigma$ and electric potential $\varphi$. The electrode contains active-material, conductive-binder, and pore phases, which form a trimodal conductivity field with a contrast of $10^{6}$. It also includes near-insulating pores and conductive structures only one to two pixels wide. Each dataset contains $10{,}000$ training and $1{,}000$ validation coefficient--solution pairs on a $64\times64$ cell-centered grid. More information is provided in the Supplementary Material.

\paragraph{Metrics.} We use four metrics in physical space. Physical Residual Fidelity (\textbf{PRF}) is the median Jacobi-preconditioned conservative residual and evaluates local physical consistency. The negative-coefficient rate (\textbf{Neg}) is the fraction of inadmissible coefficient values and evaluates positivity. The worst per-phase Wasserstein-1 distance of $\ln a$ (\textbf{$W_1$}) compares generated and reference coefficient distributions within each phase and evaluates whether even the least accurately generated phase has the correct magnitude. Interface sharpness (\textbf{Sharp}) is the Wasserstein-1 distance between the generated and reference distributions of neighboring-cell jumps in $\ln a$ and evaluates the preservation of sharp interfaces. Definitions are provided in the Supplementary Material.


\begin{figure*}[t]
    \centering
    \includegraphics[width=\textwidth]
    {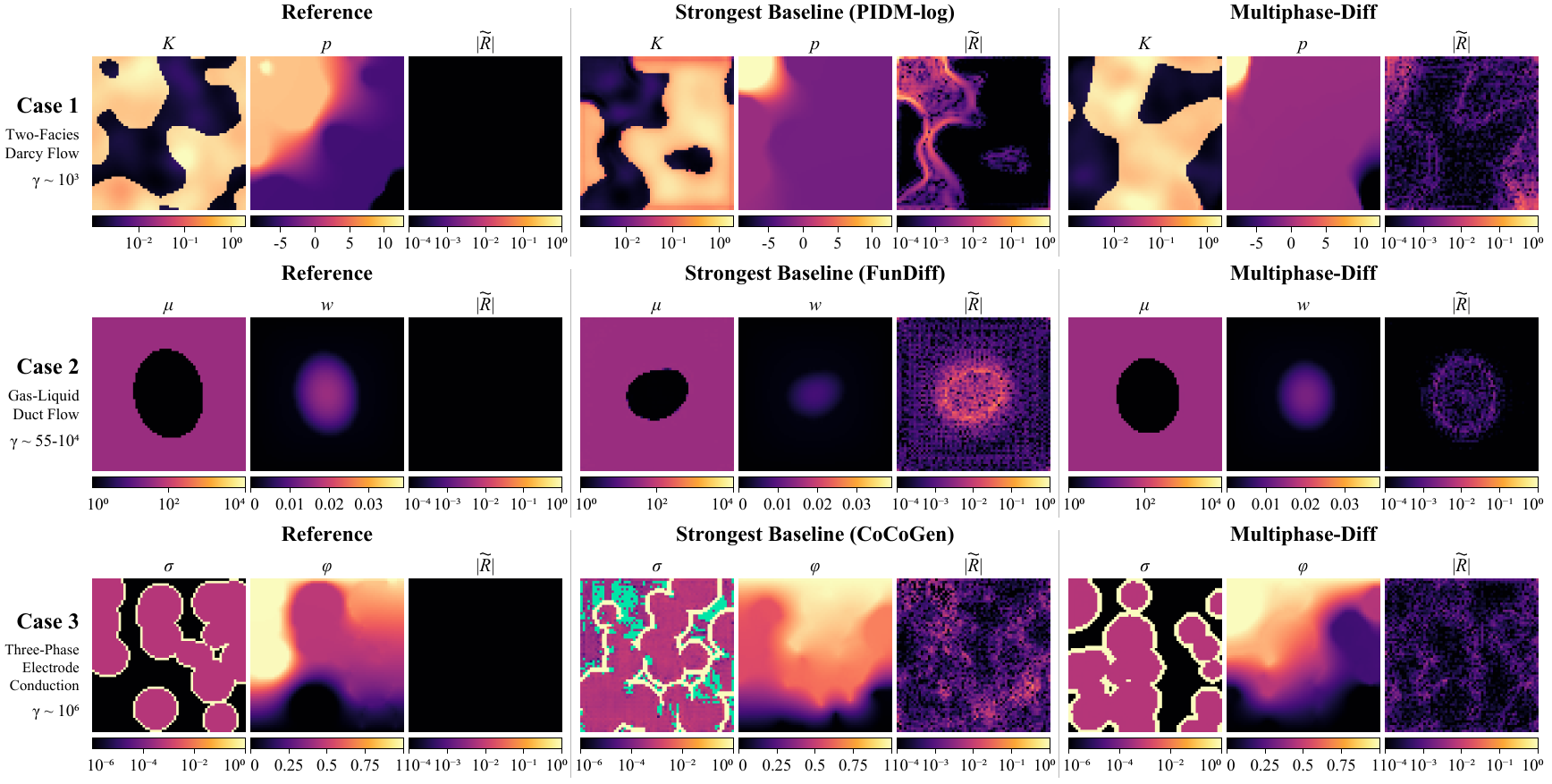}
    \caption{Qualitative comparison of coefficient, solution, and Jacobi-preconditioned residual magnitude $|\widetilde R|$ across three cases. The strongest baseline is selected by the lowest PRF; green pixels in CoCoGen's $\sigma$ field indicate non-positive conductivity.}
    \label{fig:qualitative_comparison}
\end{figure*}

\paragraph{Implementation.} All diffusion models share a $12.78$M-parameter 2D U-Net, with extra conditioning channels only for PG-Diffusion. We use $T=100$ cosine-scheduled steps and $x_0$-prediction with Min-SNR-5 weighting. Each model is trained for $100{,}000$ steps with Adam at a learning rate of $10^{-4}$. The batch size is $32$, and the EMA decay is $0.999$. Each model is evaluated on $256$ samples. The weights on the preconditioned flux and pointwise residuals are $0.1$ and $10^{-3}$. All experiments use a single NVIDIA RTX~4090 GPU.

\subsection{Baseline Comparison}
\label{sec:baseline_comparison}

We compare Multiphase-Diff with seven baselines. \textbf{DDPM}~\cite{ho2020ddpm} serves as the physics-free reference. \textbf{PG-Diffusion}~\cite{shu2023physics} conditions denoising on the PDE-residual gradient. \textbf{CoCoGen}~\cite{jacobsen2025cocogen} corrects clean estimates through residual descent during sampling, whereas \textbf{DiffusionPDE}~\cite{huang2024diffusionpde} guides noisy samples by backpropagating the physical residual through the denoiser. \textbf{FunDiff}~\cite{wang2026fundiff} learns a physics-informed function autoencoder before generating latent representations with rectified flow. \textbf{PIDM}~\cite{bastek2025pidm} incorporates a conventional pointwise residual into diffusion training. Finally, \textbf{PIDM-log} augments PIDM with the positivity-preserving logarithmic representation from Section~\ref{sec:representation}, isolating the benefit of this representation from the remaining proposed components.

Across all three benchmarks, Multiphase-Diff achieves the strongest overall balance between physical and distributional fidelity. As shown in Table~\ref{tab:baseline_comparison} and Figure~\ref{fig:metric_comparison}, it obtains the lowest PRF, worst per-phase $W_1$, and interface-sharpness error in every case while maintaining a zero negative-coefficient rate. Relative to the strongest competing method in each case, Multiphase-Diff reduces PRF by factors of $2.3$, $6.4$, and $3.0$, respectively. The four complementary metrics collectively demonstrate this overall advantage, as several baselines preserve positivity but still exhibit substantial physical-residual, phase-distribution, or interface errors. The generated coefficient--solution pairs and their local residual fields in Figure~\ref{fig:qualitative_comparison} further illustrate how Multiphase-Diff preserves sharp multiphase structure while maintaining local physical consistency, even under extreme contrast and thin phases.

The three cases highlight complementary capabilities of the proposed framework. \textbf{Case~1} shows that Multiphase-Diff recovers both the magnitude of a low-permeability phase and its irregular sharp interfaces when the principal difficulty lies in the discontinuous coefficient. \textbf{Case~2} shows that it preserves sharp gas--liquid viscosity interfaces together with their heterogeneous velocity responses, demonstrating its effectiveness when discontinuous coefficients are coupled with strongly disparate solution scales. \textbf{Case~3} provides the most stringent test, combining a conductivity contrast of $10^{6}$, three material phases, near-insulating pores, and conductive structures only one to two pixels wide. Multiphase-Diff preserves all three conductivity modes without non-positive values and achieves substantially lower phase-distribution and interface errors, demonstrating its effectiveness under extreme contrast, thin phases, and near-degeneracy.

\begin{table*}[t]
\centering
\small
\setlength{\tabcolsep}{2pt}
\begin{tabular*}{\textwidth}{
@{\extracolsep{\fill}}lccccc@{}
}
\toprule
Variant &
PRF ($\times10^{-3}$) $\downarrow$ &
Worst $W_1$ $\downarrow$ &
Sharp $\downarrow$ &
Int. &
PFE $\downarrow$ \\
\midrule

w/o Flux Res. &
$0.826/1.61/3.15$ &
$0.179/1.422/1.282$ &
$0.055/0.128/0.193$ &
$62.2/0.2/164.3$ &
$0.098/0.080/0.013$ \\

w/o Bijection &
\multicolumn{3}{c}{Diverged} &
$0/0/0$ &
$0.492/0.739/0.384$ \\

w/o Jacobi &
$\mathbf{0.085}/\mathbf{0.044}/1.05$ &
$2.021/2.647/0.982$ &
$0.137/0.166/0.213$ &
$0/0/2.4$ &
$0.316/0.098/0.032$ \\

\midrule
\textbf{Multiphase-Diff} &
$0.358/0.230/\mathbf{0.417}$ &
$\mathbf{0.008}/\mathbf{0.933}/\mathbf{0.025}$ &
$\mathbf{0.023}/\mathbf{0.023}/\mathbf{0.039}$ &
$\mathbf{63.8}/\mathbf{64.2}/\mathbf{187.9}$ &
$\mathbf{0.045}/\mathbf{0.073}/\mathbf{0.010}$ \\
\bottomrule
\end{tabular*}
\caption{Component ablations on Cases~1/2/3. Lower is better for metrics marked by $\downarrow$; Int. diagnoses interface collapse and should remain comparable to that of the complete Multiphase-Diff model. Best values are bold.}
\label{tab:ablations}
\end{table*}

\subsection{Component Ablations}
\label{sec:ablations}

We ablate the three components introduced in Section~\ref{sec:methodology}, with results reported in Table~\ref{tab:ablations}. We additionally report \textbf{interfaces per sample (Int.)} to detect interface collapse and \textbf{phase-fraction error (PFE)} to measure composition errors.

\paragraph{w/o Conservative Flux Residual.}
The \textbf{w/o Flux Res.} variant replaces the conservative flux residual with the pointwise residual, yielding PIDM-log. PRF increases across all three cases; moreover, Int. decreases from $64.2$ to $0.2$ in Case~2, while the worst $W_1$ increases from $0.025$ to $1.282$ in Case~3. These results show that the conservative residual preserves sharp and thin phases while enforcing local conservation.

\paragraph{w/o Analytic Bijective Representation.}
The \textbf{w/o Bijection} variant replaces the analytic bijection with affine normalization, causing all runs to diverge at steps $300$, $700$, and $1100$. Affine decoding produces non-positive coefficients and destabilizes harmonic transmissibilities, whereas the bijection ensures positivity and stable flux-based training.

\paragraph{w/o Jacobi Preconditioning.}
The \textbf{w/o Jacobi} variant trains with the raw residual, whose uneven gradients overwhelm the data objective and encourage the model to shrink coefficient contrast and flatten the solution. The resulting interface-free fields are easier to satisfy physically, explaining the deceptively lower PRF in Cases~1 and 2 despite zero interfaces and worst $W_1$ values of $2.021$ and $2.647$. In Case~3, the raw residual underweights the near-insulating phase, whose errors instead increase the preconditioned PRF. Jacobi preconditioning prevents both failures by balancing physics gradients across phases and samples.

\subsection{Robustness Study}
\label{sec:robustness}

To further assess the robustness of Multiphase-Diff for high-contrast, sharp-interface multiphase systems, we examine its sensitivity to phase contrast and composition through two controlled five-point sweeps based on Case~1. The contrast sweep fixes the high-permeability volume fraction at $v_f=0.5$ and varies $\gamma\in\{10^1,10^2,10^3,10^4,10^5\}$, while the composition sweep fixes $\gamma=10^3$ and varies $v_f\in\{0.1,0.3,0.5,0.7,0.9\}$, including settings in which either phase becomes a minority. Multiphase-Diff is trained independently at each setting and evaluated using the same four metrics as in the main comparison.

\begin{figure}[t]
    \centering
    \includegraphics[width=\linewidth]{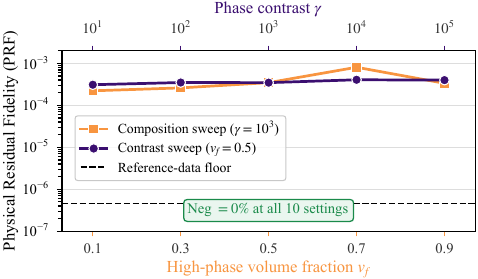}
    \caption{Robustness study to phase contrast and composition. The upper and lower axes show the two sweeps; PRF remains of order $10^{-4}$ and Neg is zero. The dashed line marks the reference-data floor.}
    \label{fig:robustness}
\end{figure}

As shown in Figure~\ref{fig:robustness}, PRF remains within $3.10\times10^{-4}$--$4.07\times10^{-4}$ across the contrast sweep and $2.22\times10^{-4}$--$8.09\times10^{-4}$ across the composition sweep, with Neg remaining zero throughout. The worst per-phase $W_1$ stays below $0.034$ and $0.018$, respectively, while Sharp remains below $0.021$ and $0.010$. The largest composition-sweep PRF occurs at $v_f=0.7$ rather than at either composition extreme. These results show that Multiphase-Diff maintains physical consistency, phase fidelity, and sharp interfaces across four orders of magnitude in contrast and retains minority phases occupying only about $10\%$ of the domain.

\section{Conclusion}

We presented \textbf{Multiphase-Diff}, a physics-constrained diffusion framework for generative modeling of high-contrast, sharp-interface multiphase fields. To overcome the interface irregularity, low-amplitude phase under-resolution, and residual-scale heterogeneity that arise specifically in this regime, Multiphase-Diff makes three corresponding methodological contributions: a conservative flux residual, an analytic bijective representation, and a Jacobi-preconditioned likelihood, respectively. Extensive experiments on three complementary benchmarks, including comparisons with seven baselines, component ablations, and controlled robustness sweeps, validate the roles of these mechanisms and the framework's robustness across phase contrasts and compositions. In the three cases, Multiphase-Diff reduces PRF by factors of $2.3$, $6.4$, and $3.0$, respectively, over the strongest baseline while preserving coefficient positivity, per-phase distributions, and sharp interfaces.

\noindent\textbf{Limitations.} Our current study is limited to steady elliptic conservation laws. Future work will extend Multiphase-Diff to time-dependent systems with evolving interfaces, which will require space--time conservative residuals and temporally consistent generative modeling.

\bibliography{reference.bib}

\setcounter{secnumdepth}{2} 

\twocolumn[
\begin{center}
    {\Large\bfseries Supplementary Materials}
\end{center}
\medskip
]

\section{Equivalence of Clean-Sample and Noise Prediction}
\label{sec:equivalence}

This section supplies the timestep-dependent equivalence stated in
Section~4.4 of the main paper.  In latent space, the forward diffusion
process is
\begin{equation}
  z_t=\sqrt{\abar_t}\,z_0+\sqrt{1-\abar_t}\,\epsilon,
  \qquad \epsilon\sim\Normal(0,I).
  \label{eq:forward}
\end{equation}
Given a noise predictor $\epsilon_\theta(z_t,t)$, the corresponding
clean-sample prediction is
\begin{equation}
  \widehat z_{0,\theta}(z_t,t)
  =
  \frac{z_t-\sqrt{1-\abar_t}\,
  \epsilon_\theta(z_t,t)}
  {\sqrt{\abar_t}}.
  \label{eq:x0_from_eps}
\end{equation}
Conversely, a clean-sample predictor determines a noise prediction through
\begin{equation}
  \epsilon_\theta(z_t,t)
  =
  \frac{z_t-\sqrt{\abar_t}\,
  \widehat z_{0,\theta}(z_t,t)}
  {\sqrt{1-\abar_t}}.
  \label{eq:eps_from_x0}
\end{equation}
Subtracting \eqref{eq:eps_from_x0} from the true noise obtained from
\eqref{eq:forward} gives
\begin{equation}
  \epsilon-\epsilon_\theta(z_t,t)
  =
  \sqrt{\frac{\abar_t}{1-\abar_t}}\,
  \bigl(\widehat z_{0,\theta}(z_t,t)-z_0\bigr).
  \label{eq:error_relation}
\end{equation}
Therefore,
\begin{equation}
\begin{aligned}
  &\left\|\epsilon-\epsilon_\theta(z_t,t)\right\|_2^2
  \\
  &\quad=
  \operatorname{SNR}(t)
  \left\|z_0-\widehat z_{0,\theta}(z_t,t)\right\|_2^2,
  \\
  &\operatorname{SNR}(t)=\frac{\abar_t}{1-\abar_t}.
\end{aligned}
  \label{eq:loss_relation}
\end{equation}
Thus, a weighted noise-prediction objective
\begin{equation}
  \E\!\left[
    w_t^{(\epsilon)}
    \left\|\epsilon-\epsilon_\theta(z_t,t)\right\|_2^2
  \right]
\end{equation}
is exactly a clean-sample regression objective with
\begin{equation}
  w_t^{(z_0)}
  =
  w_t^{(\epsilon)}\operatorname{SNR}(t).
  \label{eq:weight_conversion}
\end{equation}

For Min-SNR-$\gamma$ weighting, the noise-prediction weight is
\begin{equation}
  w_t^{(\epsilon)}
  =
  \frac{\min\{\operatorname{SNR}(t),\gamma\}}
       {\operatorname{SNR}(t)}.
\end{equation}
Equation~\eqref{eq:weight_conversion} therefore yields the clean-sample
weight
\begin{equation}
  \lambda_t
  =
  w_t^{(z_0)}
  =
  \min\{\operatorname{SNR}(t),\gamma\}.
  \label{eq:minsnr_x0}
\end{equation}
We use $\gamma=5$.  The latent data term used in the main paper is hence
\begin{equation}
  \mathcal{L}_{\mathrm{data}}
  =
  \E_{t,z_0,\epsilon}\!\left[
    \lambda_t
    \left\|z_0-\widehat z_{0,\theta}(z_t,t)\right\|_2^2
  \right].
\end{equation}
The equivalence above concerns the diffusion regression term.  Physics is
evaluated on the decoded clean estimate
$g^{-1}(\widehat z_{0,\theta})$, as specified in Section~4.4 of the main
paper; it is not evaluated on the noisy latent $z_t$.

\section{Benchmark and Dataset Details}
\label{sec:datasets}

\subsection{Unified governing problem and discretization}

All three benchmarks generate coefficient--solution pairs $(a,u)$ governed
by
\begin{equation}
  \nabla\!\cdot\!\left(a(\boldsymbol{x})\nabla
  u(\boldsymbol{x})\right)=f(\boldsymbol{x}),
  \qquad \boldsymbol{x}\in\Omega=[0,1]^2.
  \label{eq:elliptic}
\end{equation}
The coefficient is positive and piecewise constant or nearly piecewise
constant.  Across an internal material interface $\Gamma$, the weak
solution satisfies
\begin{equation}
  [u]_\Gamma=0,
  \qquad
  [a\nabla u\cdot\boldsymbol{n}]_\Gamma=0.
  \label{eq:interface_conditions}
\end{equation}
Consequently, $u$ remains continuous while its normal derivative can jump
by the coefficient ratio.

We discretize \eqref{eq:elliptic} on a $64\times64$ cell-centered grid with
spacing $H=1/64$.  For neighboring cells $i$ and $j$, the two-point
transmissibility is the harmonic mean
\begin{equation}
  T_{ij}=\frac{2a_i a_j}{a_i+a_j}.
  \label{eq:harmonic}
\end{equation}
Writing $\mathcal{N}(i)$ for the set of cell neighbors and
$\mathcal{B}_D(i)$ for Dirichlet boundary faces adjacent to cell $i$, the
discrete conservative residual is
\begin{equation}
\begin{aligned}
  R_i(a,u,f)
  &=
  \frac{1}{H^2}
  \sum_{j\in\mathcal{N}(i)}
  T_{ij}(u_j-u_i)
  \\
  &\quad+
  \frac{1}{H^2}
  \sum_{b\in\mathcal{B}_D(i)}
  T_{ib}(u_b-u_i)
  -\bar f_i.
\end{aligned}
  \label{eq:fv_residual}
\end{equation}
Prescribed Neumann fluxes are absorbed into the boundary-adjusted source
$\bar f_i$.  The same face fluxes, source discretization, and boundary
treatment are used by the data-generating solver, the training residual,
and evaluation.  Each benchmark contains 10,000 training pairs and 1,000
validation pairs generated with disjoint random seeds.  Phase labels are
retained for evaluation only and are not supplied to the unconditional
generators.

\subsection{Case 1: two-facies Darcy flow}

The generated pair is permeability and pressure,
\begin{equation}
  (a,u)=(K,p),\qquad
  \nabla\!\cdot(K\nabla p)=-f_s.
  \label{eq:darcy}
\end{equation}
The permeability contains irregular, sharp two-facies interfaces with the
characteristic contrast of approximately $10^3$ reported in the main paper.
Because permeability varies continuously within each facies, the empirical
global maximum-to-minimum ratio over the validation split is
$6.73\times10^3$.  A fixed dipole forcing drives the flow: $f_s=+10$ in the
top-left corner region covering $12.5\%$ of each coordinate extent and
$f_s=-10$ in the corresponding bottom-right region.  All external
boundaries are impermeable,
$K\nabla p\cdot\boldsymbol{n}=0$, and the additive pressure nullspace is
removed using the zero-mean gauge
\begin{equation}
  \sum_i p_i=0.
\end{equation}
The source and sink have equal total strength, ensuring compatibility with
the Neumann problem.  The benchmark jointly tests sharp permeability jumps
and the corresponding discontinuity of the pressure gradient.

\subsection{Case 2: fully developed gas--liquid duct flow}

The generated pair is viscosity and axial velocity,
\begin{equation}
  (a,u)=(\mu,w),\qquad
  \nabla\!\cdot(\mu\nabla w)=-G,
  \label{eq:duct}
\end{equation}
where $G=1$ is the imposed axial pressure-gradient magnitude.  Fully
developed flow reduces the axial momentum equation to the cross-sectional
elliptic problem \eqref{eq:duct}.  No-slip conditions are imposed on the
duct walls,
\begin{equation}
  w=0\qquad\text{on }\partial\Omega.
\end{equation}
The two phases form sharp viscosity interfaces.  Across the interface,
velocity and tangential traction are continuous:
\begin{equation}
  [w]_\Gamma=0,\qquad
  [\mu\nabla w\cdot\boldsymbol{n}]_\Gamma=0.
\end{equation}
The coefficient takes values in $\{1,55,10^3,10^4\}$: the gas viscosity is
normalized to one and the three liquid types yield contrasts from $55$ to
$10^4$.  The validation phase fractions are $0.26/0.74$ for gas/liquid.
Together with the heterogeneous velocity scales, this makes both the
coefficient and solution channels multiscale.

\subsection{Case 3: three-phase electrode conduction}

The generated pair is electronic conductivity and electric potential,
\begin{equation}
  (a,u)=(\sigma,\varphi),\qquad
  \nabla\!\cdot(\sigma\nabla\varphi)=0.
  \label{eq:electrode}
\end{equation}
The microstructure contains pore, active-material, and conductive-binder
phases with conductivity levels $\{10^{-6},10^{-3},1\}$, respectively.
Their trimodal field spans a total contrast of $10^6$, with a near-insulating
pore phase.  We impose $\varphi=1$ and $\varphi=0$ on the $x=0$ and $x=1$
sides, respectively, and homogeneous Neumann conditions on the two remaining
sides.  The validation phase fractions are $0.42/0.47/0.11$.  The dataset includes
conductive features only one to two cells wide.  This case combines
trimodality, thin structures, and a nearly degenerate coefficient while
retaining strict positivity.

\subsection{Jacobi-preconditioned evaluation residual}

For completeness, the residual used by PRF is recalled here because it is
needed to make the metric definition unambiguous.  Define the unscaled
diagonal magnitude
\begin{equation}
  D_i=
  \sum_{j\in\mathcal{N}(i)}T_{ij}
  +
  \sum_{b\in\mathcal{B}_D(i)}T_{ib}.
\end{equation}
Let $s_{\widehat u}>0$ be the detached robust per-sample scale of the decoded
solution after fixing the solution gauge in Neumann cases.  The local
normalizer and dimensionless residual are
\begin{align}
  Q_i &=
  \frac{D_i s_{\widehat u}}{H^2}
  +\|f\|_\infty+\varepsilon,\\
  S_i&=\stopgrad(Q_i),\qquad
  \Rtilde_i=\frac{R_i}{S_i},
  \label{eq:preconditioned_residual}
\end{align}
where $\varepsilon>0$ is a numerical floor.

\section{Exact Evaluation Metrics}
\label{sec:metrics}

All metrics are computed after decoding to physical space.  Let
$\{(\widehat a^{(n)},\widehat u^{(n)})\}_{n=1}^{N_g}$ denote generated
samples and let $\{a_{\mathrm{ref}}^{(m)}\}_{m=1}^{N_r}$ denote reference
validation coefficients.  The reported experiments use $N_g=256$.

\subsection{Physical Residual Fidelity}

For each generated sample, we first average the absolute
Jacobi-preconditioned conservative residual over its cells and then take the
median across samples:
\begin{equation}
  \operatorname{PRF}
  =
  \median_{1\leq n\leq N_g}
  \left[
    \frac{1}{N_c}\sum_{i=1}^{N_c}
    \left|
      \Rtilde_i\!\left(
        \widehat a^{(n)},\widehat u^{(n)},f
      \right)
    \right|
  \right].
  \label{eq:prf}
\end{equation}
The index $i$ includes the cells whose balance equations contain boundary
contributions.  Lower values indicate better local conservation.  Scoring
held-out reference coefficient--solution pairs with the same operator gives
the reference-data floor.

\subsection{Negative-coefficient rate}

The negative-coefficient rate is the fraction of generated coefficient
values outside the admissible positive domain:
\begin{equation}
  \operatorname{Neg}
  =
  \frac{1}{N_gN_c}
  \sum_{n=1}^{N_g}\sum_{i=1}^{N_c}
  \ind\!\left[\widehat a_i^{(n)}\leq0\right],
  \qquad N_c=64^2.
  \label{eq:neg}
\end{equation}
No clipping is applied before computing this metric.

\subsection{Worst per-phase Wasserstein distance}

Phases are segmented by fixed midpoint thresholds in $\log_{10}a$.  The
threshold sets used in evaluation are
\begin{equation}
\begin{aligned}
  \mathcal{E}_1
  &=\left\{\tfrac{1}{2}\log_{10}(2\times10^{-3})\right\},\\
  \mathcal{E}_2
  &=\left\{\tfrac{1}{2}\log_{10}(55)\right\},\\
  \mathcal{E}_3&=\{-4.5,-1.5\}.
\end{aligned}
  \label{eq:phase_thresholds}
\end{equation}
These give two phases in Cases 1 and 2 and three phases in Case 3.  For
phase $k$, let $\mathcal{A}^{\mathrm{gen}}_k$ and
$\mathcal{A}^{\mathrm{ref}}_k$ be the pooled generated and reference
coefficient values assigned to that phase.  Generated values with
$\widehat a\leq0$ are excluded from the logarithmic distance and are instead
accounted for by \eqref{eq:neg}.  The phase-wise distance is
\begin{equation}
  d_k
  =
  \Wone\!\left(
    \{\ln a:a\in\mathcal{A}^{\mathrm{gen}}_k,\ a>0\},
    \{\ln a:a\in\mathcal{A}^{\mathrm{ref}}_k\}
  \right),
  \label{eq:phase_w1}
\end{equation}
and the reported statistic is
\begin{equation}
  \operatorname{WorstPhase}\text{-}\Wone=\max_k d_k.
  \label{eq:worst_w1}
\end{equation}
Using the natural logarithm makes this metric measure multiplicative
coefficient error without allowing a well-reproduced high-magnitude phase
to hide failure on a low-magnitude phase.

\subsection{Interface sharpness}

For a positive coefficient field $a$, define the multiset of horizontal
and vertical neighboring-cell jumps
\begin{equation}
  \mathcal{J}(a)
  =
  \left\{
    \left|\ln a_i-\ln a_j\right|
    : (i,j)\in\mathcal{F}_{\mathrm{int}}
  \right\},
  \label{eq:jumps}
\end{equation}
where $\mathcal{F}_{\mathrm{int}}$ contains every horizontal and vertical
interior face.  A generated face is included only if both adjacent
coefficients are positive; non-positive mass is measured separately by
Neg.  Pooling faces across samples, interface sharpness is
\begin{equation}
  \operatorname{Sharp}
  =
  \Wone\!\left(
    \mathcal{J}(\widehat a),
    \mathcal{J}(a_{\mathrm{ref}})
  \right).
  \label{eq:sharp}
\end{equation}
The distribution includes both within-phase and inter-phase neighboring
jumps.  Consequently, the statistic detects both smeared interfaces and
spurious high-frequency jumps without requiring a method-specific interface
threshold.

\section{Complete Robustness Values}
\label{sec:robustness}

Table~\ref{tab:robustness} lists the numerical values underlying the two
robustness curves in Section~5.4 of the main paper.  These are controlled
Darcy benchmarks: the composition is fixed at $v_f=0.5$ in the contrast
sweep, and the contrast is fixed at $\gamma=10^3$ in the composition sweep.
Lower is better for all metrics.

\begin{table}[t]
  \centering
  \small
  \begin{tabular}{lrrrr}
    \toprule
    Setting & PRF & Neg (\%) & Worst $\Wone$ & Sharp\\
    \midrule
    \multicolumn{5}{l}{\emph{Contrast sweep, $v_f=0.5$}}\\
    $\gamma=10^{1}$ & $3.10{\times}10^{-4}$ & 0.0 & 0.009 & 0.006\\
    $\gamma=10^{2}$ & $3.50{\times}10^{-4}$ & 0.0 & 0.015 & 0.009\\
    $\gamma=10^{3}$ & $3.48{\times}10^{-4}$ & 0.0 & 0.013 & 0.009\\
    $\gamma=10^{4}$ & $4.07{\times}10^{-4}$ & 0.0 & 0.034 & 0.016\\
    $\gamma=10^{5}$ & $3.99{\times}10^{-4}$ & 0.0 & 0.029 & 0.021\\
    \midrule
    \multicolumn{5}{l}{\emph{Composition sweep, $\gamma=10^3$}}\\
    $v_f=0.1$ & $2.22{\times}10^{-4}$ & 0.0 & 0.009 & 0.006\\
    $v_f=0.3$ & $2.61{\times}10^{-4}$ & 0.0 & 0.006 & 0.005\\
    $v_f=0.5$ & $3.48{\times}10^{-4}$ & 0.0 & 0.013 & 0.009\\
    $v_f=0.7$ & $8.09{\times}10^{-4}$ & 0.0 & 0.018 & 0.008\\
    $v_f=0.9$ & $3.35{\times}10^{-4}$ & 0.0 & 0.018 & 0.007\\
    \bottomrule
  \end{tabular}
  \caption{Complete results for the controlled contrast and composition
  sweeps reported in the main paper.}
  \label{tab:robustness}
\end{table}

\section{Baseline Implementations and Complete Results}
\label{sec:complete_results}

\subsection{Baseline implementation details}

All diffusion methods except FunDiff use the shared protocol reported in the
main paper: a 12.78M-parameter U-Net, $T=100$ cosine-scheduled steps,
$x_0$-prediction with Min-SNR-5 weighting, Adam with learning rate
$10^{-4}$, batch size 32, EMA decay 0.999, 100,000 training steps, and 256
evaluation samples.  FunDiff instead uses a function autoencoder with eight
latent channels followed by rectified flow (40k + 60k steps), as listed in
Table~\ref{tab:baseline_implementation}.  The table records the
method-specific settings used in the final runs.  PG-Diffusion uses two
additional conditioning channels, null-condition dropout 0.1, and guidance
scale 3.0.  CoCoGen and DiffusionPDE reuse the DDPM checkpoint and introduce
physics only during sampling.

\begin{table*}[t]
\centering
\scriptsize
\begin{tabular}{lp{0.95in}p{1.15in}p{1.25in}cp{2.05in}}
\toprule
Method & Physics location & Representation & Residual & Weight & Training or sampling modification\\
\midrule
DDPM & none & affine standardization & none & 0 & standard reverse diffusion\\
PG-Diffusion & conditioning & affine standardization & pointwise strong form & $10^{-3}$ & two conditioning channels; guidance scale 3.0\\
CoCoGen & sampling correction & affine standardization & flux + Jacobi at sampling & 0 & $N=100$, $M=10$, $\eta=0.05$\\
DiffusionPDE & sampling guidance & affine standardization & flux + Jacobi at sampling & 0 & $\zeta=0.1$ over all $T$ steps\\
FunDiff & autoencoder prior & function autoencoder, 8 latent channels & physics-informed autoencoder & 0.1 & FAE 40k steps + rectified flow 60k steps\\
PIDM & training likelihood & affine standardization & pointwise strong form & $10^{-3}$ & standard reverse diffusion\\
PIDM-log & training likelihood & bijective log/asinh & pointwise strong form & $10^{-3}$ & standard reverse diffusion\\
\textbf{Multiphase-Diff} & training likelihood & bijective log/asinh & conservative flux + Jacobi & 0.1 & standard reverse diffusion; no sampling correction\\
\bottomrule
\end{tabular}
\caption{Implementation settings for the seven baselines and
Multiphase-Diff.  Weights denote the training-time physics coefficient;
inference-time methods reuse the physics-free DDPM checkpoint.}
\label{tab:baseline_implementation}
\end{table*}

\subsection{Complete per-phase distribution distances}

Table~\ref{tab:per_phase_w1} expands the worst-phase statistic in the main
paper into its individual phase contributions.  Each row maximum reproduces
the corresponding worst per-phase $W_1$ in the main-paper Table~1.

\begin{table*}[t]
\centering
\small
\begin{tabular}{lrrrrrrr}
\toprule
& \multicolumn{2}{c}{Case 1} & \multicolumn{2}{c}{Case 2} & \multicolumn{3}{c}{Case 3}\\
\cmidrule(lr){2-3}\cmidrule(lr){4-5}\cmidrule(lr){6-8}
Method & low $K$ & high $K$ & gas & liquid & pore & active & binder\\
\midrule
DDPM              & 0.144 & 0.006 & 1.141 & 0.538 & 2.522 & 0.682 & 0.001\\
PG-Diffusion      & 0.298 & 0.005 & 1.140 & 0.667 & 2.520 & 0.725 & 0.001\\
CoCoGen           & 0.116 & 0.005 & 1.149 & 0.580 & 2.715 & 0.508 & 0.001\\
DiffusionPDE      & 0.174 & 0.006 & 1.119 & 1.187 & 2.502 & 0.681 & 0.001\\
FunDiff           & 0.038 & 0.033 & 0.051 & 1.339 & 0.146 & 0.219 & 0.739\\
PIDM              & 1.305 & 0.145 & 1.884 & 3.397 & 2.486 & 1.047 & 1.228\\
PIDM-log          & 0.028 & 0.179 & 0.248 & 1.422 & 0.059 & 0.080 & 1.282\\
\textbf{Multiphase-Diff}
                  & 0.008 & 0.004
                  & 0.010 & 0.933
                  & 0.013 & 0.009 & 0.025\\
\bottomrule
\end{tabular}
\caption{Per-phase Wasserstein-1 distance of $\ln a$ on the positive
support.  Lower is better.  Non-positive coefficient values are excluded
from this logarithmic distance and reported separately by Neg.}
\label{tab:per_phase_w1}
\end{table*}

\subsection{Additional diagnostics}

Table~\ref{tab:diagnostics} reports quantities computed in the main
evaluation but omitted from the main-paper comparison for space.  PRF mean
is more sensitive than its median counterpart to high-residual samples.
Phase-fraction error (PFE) is the mean absolute error in phase area
fractions.  The interface count is the number of coefficient interfaces per
sample satisfying the common detection criterion; it is a diagnostic for
interface collapse rather than a monotone score to maximize.

\begin{table*}[t]
\centering
\scriptsize
\begin{tabular}{clrrrrr}
\toprule
Case & Method & PRF median & PRF mean & PFE & Interfaces/sample & Neg (\%)\\
\midrule
1 & DDPM             & $1.09\times10^{-2}$ & $2.56\times10^{-2}$ & 0.005 & 29.0 & 1.9\\
1 & PG-Diffusion     & $1.34\times10^{-2}$ & $3.98\times10^{-2}$ & 0.011 & 40.7 & 4.5\\
1 & CoCoGen          & $1.96\times10^{-2}$ & $1.83\times10^{-1}$ & 0.000 & 49.0 & 6.9\\
1 & DiffusionPDE     & $9.76\times10^{-3}$ & $4.77\times10^{-2}$ & 0.052 & 29.9 & 2.8\\
1 & FunDiff          & $6.28\times10^{-3}$ & $7.26\times10^{-3}$ & 0.025 & 93.1 & 0.0\\
1 & PIDM             & $9.81\times10^{-4}$ & $2.93\times10^{-3}$ & 0.360 & 9.2 & 2.4\\
1 & PIDM-log         & $8.26\times10^{-4}$ & $2.39\times10^{-3}$ & 0.117 & 65.4 & 0.0\\
1 & \textbf{Multiphase-Diff} & $3.58\times10^{-4}$ & $5.40\times10^{-4}$ & 0.045 & 63.8 & 0.0\\
\midrule
2 & DDPM             & $8.57\times10^{-2}$ & $1.04\times10^{-1}$ & 0.090 & 44.4 & 7.0\\
2 & PG-Diffusion     & $1.14\times10^{-1}$ & $1.45\times10^{-1}$ & 0.089 & 51.2 & 11.1\\
2 & CoCoGen          & $1.60\times10^{-1}$ & $1.81\times10^{-1}$ & 0.077 & 55.9 & 14.8\\
2 & DiffusionPDE     & $9.18\times10^{-2}$ & $1.10\times10^{-1}$ & 0.077 & 47.2 & 8.8\\
2 & FunDiff          & $1.47\times10^{-3}$ & $2.17\times10^{-3}$ & 0.081 & 51.7 & 0.0\\
2 & PIDM             & $4.57\times10^{-1}$ & $4.57\times10^{-1}$ & 0.207 & 0.0 & 43.0\\
2 & PIDM-log         & $1.61\times10^{-3}$ & $7.52\times10^{-3}$ & 0.068 & 0.3 & 0.0\\
2 & \textbf{Multiphase-Diff} & $2.30\times10^{-4}$ & $3.40\times10^{-4}$ & 0.073 & 64.2 & 0.0\\
\midrule
3 & DDPM             & $3.43\times10^{-3}$ & $3.76\times10^{-3}$ & 0.144 & 133.4 & 20.0\\
3 & PG-Diffusion     & $3.16\times10^{-3}$ & $3.65\times10^{-3}$ & 0.090 & 126.8 & 28.1\\
3 & CoCoGen          & $1.24\times10^{-3}$ & $1.35\times10^{-3}$ & 0.067 & 132.5 & 33.1\\
3 & DiffusionPDE     & $3.23\times10^{-3}$ & $3.66\times10^{-3}$ & 0.102 & 121.1 & 27.0\\
3 & FunDiff          & $4.48\times10^{-3}$ & $4.50\times10^{-3}$ & 0.025 & 102.7 & 0.0\\
3 & PIDM             & $1.68\times10^{-3}$ & $1.75\times10^{-3}$ & 0.260 & 47.7 & 3.2\\
3 & PIDM-log         & $3.15\times10^{-3}$ & $3.40\times10^{-3}$ & 0.013 & 159.8 & 0.0\\
3 & \textbf{Multiphase-Diff} & $4.17\times10^{-4}$ & $4.51\times10^{-4}$ & 0.010 & 187.9 & 0.0\\
\bottomrule
\end{tabular}
\caption{Additional diagnostics for all eight methods.  Lower is better for
PRF, PFE, and Neg.  Interface counts diagnose preservation or collapse and
should be interpreted relative to the reference morphology.}
\label{tab:diagnostics}
\end{table*}

\section{Reference-Data Floors}
\label{sec:floors}

To quantify finite-sample and numerical floors, we score the held-out
validation pairs with the same residual and compare five fixed random
split-half partitions (seeds 0--4) for the two distributional metrics.
Table~\ref{tab:floors} reports the validation PRF median and Neg rate, and
the split-half mean $\pm$ standard deviation for worst-phase $W_1$ and
Sharp.

\begin{table*}[t]
\centering
\small
\begin{tabular}{lrrrr}
\toprule
Case & PRF & Neg (\%) & Worst $W_1$ & Sharp\\
\midrule
Darcy & $4.57\times10^{-7}$ & 0.0 & $0.0030\pm0.0010$ & $0.0069\pm0.0031$\\
Gas--liquid & $3.69\times10^{-7}$ & 0.0 & $0.1406\pm0.0799$ & $0.0034\pm0.0017$\\
Electrode & $2.17\times10^{-7}$ & 0.0 & $0.0000\pm0.0000$ & $0.0066\pm0.0038$\\
\bottomrule
\end{tabular}
\caption{Reference-data floors under the exact evaluation conventions of
the main paper.}
\label{tab:floors}
\end{table*}

Case 1 has a small nonzero $W_1$ floor because permeability varies
continuously within each facies.  Case 2 uses $\mu\in\{1,55,10^3,10^4\}$;
the liquid-phase floor reflects variation in the finite-sample mixture of
the three liquid viscosities.  Case 3 uses exactly three conductivity
levels, so each phase is a point mass and the per-phase $W_1$ floor is zero.

\section{Additional Qualitative Samples}
\label{sec:qualitative}

\begin{figure*}[t]
\centering
\includegraphics[width=\textwidth]{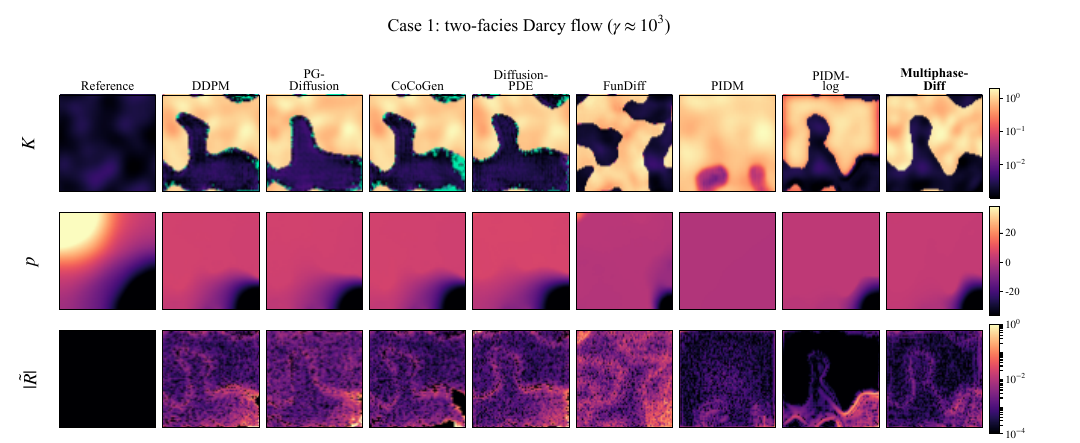}
\caption{Qualitative comparison of the reference data, all seven baselines,
and Multiphase-Diff on Case 1.  Columns correspond to methods; rows show
permeability $K$, pressure $p$, and Jacobi-preconditioned conservative
residual magnitude $|\widetilde R|$; green denotes non-positive permeability.}
\label{fig:qual_case1}
\end{figure*}

\begin{figure*}[t]
\centering
\includegraphics[width=\textwidth]{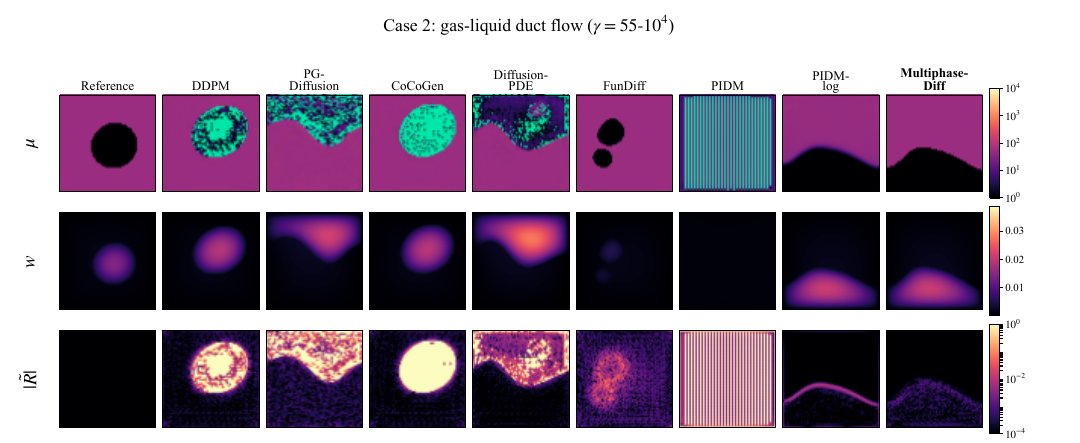}
\caption{Qualitative comparison of the reference data, all seven baselines,
and Multiphase-Diff on Case 2.  Rows show viscosity $\mu$, axial velocity
$w$, and Jacobi-preconditioned conservative residual magnitude
$|\widetilde R|$; green denotes non-positive viscosity.}
\label{fig:qual_case2}
\end{figure*}

\begin{figure*}[t]
\centering
\includegraphics[width=\textwidth]{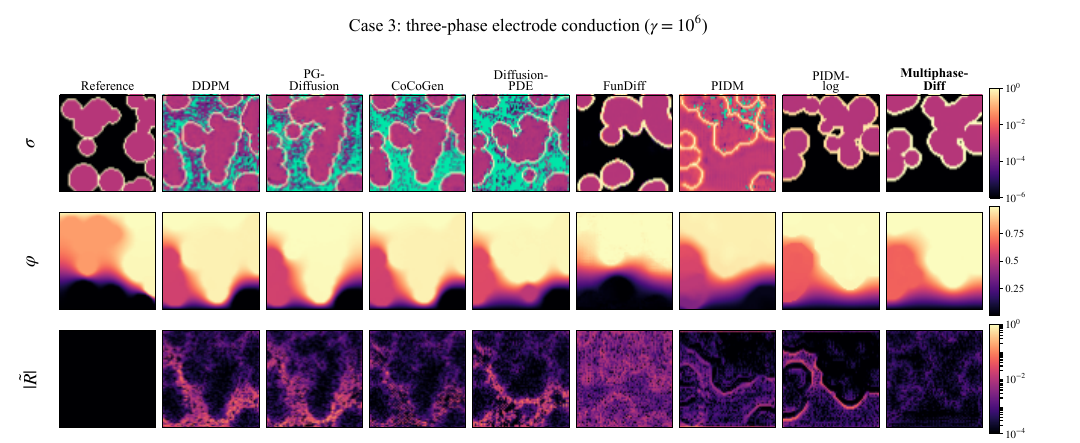}
\caption{Qualitative comparison of the reference data, all seven baselines,
and Multiphase-Diff on Case 3.  Rows show conductivity $\sigma$, electric
potential $\varphi$, and Jacobi-preconditioned conservative residual
magnitude $|\widetilde R|$; green denotes non-positive conductivity.}
\label{fig:qual_case3}
\end{figure*}

Figures~\ref{fig:qual_case1}--\ref{fig:qual_case3} compare the reference
data, all seven baselines, and Multiphase-Diff.  A fixed array index (0) was
selected before visual inspection and used for every method.  Because the
models are unconditional, equal array indices do not imply paired samples;
the fixed index provides only a reproducible, non-cherry-picked selection.
Each column contains the coefficient, solution, and preconditioned residual
magnitude.  Color scales are shared within each case, and green marks
non-positive generated coefficients.

\end{document}